\documentclass[sigconf,natbib=true,anonymous=false]{acmart}
\usepackage{multirow}
\usepackage{subfigure}
\AtBeginDocument{%
  \providecommand\BibTeX{{%
    \normalfont B\kern-0.5em{\scshape i\kern-0.25em b}\kern-0.8em\TeX}}}

\setcopyright{acmlicensed}
\usepackage{amsmath}

\newcommand{\textFunction}[1]{\text{\textbf{\textsc{text}}}\left(#1\right)}

\copyrightyear{2014}
\acmYear{2025}
\acmDOI{XXXXXXX.XXXXXXX}

\acmConference[The 48'th ACM SIGIR conference]{}{July 13th to 18th, 2025}{Padua, Italy}
\acmISBN{978-1-4503-XXXX-X/18/06}

\begin{document}

\title[Hierarchical Classification]{Hierarchical Job Classification with Similarity Graph Integration}

%\shorttitle{Forecasting Application Counts}

\author{
    Md Ahsanul Kabir$^1$, Kareem Abdelfatah$^1$, Mohammed Korayem$^1$, Mohammad Al Hasan$^2$\\
    \textnormal{$^1$CareerBuilder LLC},
    \textnormal{$^2$Indiana University Indianapolis, USA}\\
    \textnormal{\{mdahsanul.kabir, kareem.abdelfatah, mohammed.korayem\}@careerbuilder.com, alhasan@iu.edu}
}

\renewcommand{\shortauthors}{kabir, et al.}

\newcommand{\name}{{\sc HierarchicalClassification}}

\begin{abstract}

In the dynamic realm of online recruitment, accurate job classification is paramount for optimizing job recommendation systems, search rankings, and labor market analyses. As job markets evolve, the increasing complexity of job titles and descriptions necessitates sophisticated models that can effectively leverage intricate relationships within job data. Traditional text classification methods often fall short, particularly due to their inability to fully utilize the hierarchical nature of industry categories. To address these limitations, we propose a novel representation learning and classification model that embeds jobs and hierarchical industry categories into a latent embedding space. Our model integrates the Standard Occupational Classification (SOC) system and an in-house hierarchical taxonomy, Carotene, to capture both graph and hierarchical relationships, thereby improving classification accuracy. By embedding hierarchical industry categories into a shared latent space, we tackle cold start issues and enhance the dynamic matching of candidates to job opportunities. Extensive experimentation on a large-scale dataset of job postings demonstrates the model's superior ability to leverage hierarchical structures and rich semantic features, significantly outperforming existing methods. This research provides a robust framework for improving job classification accuracy, supporting more informed decision-making in the recruitment industry.

\end{abstract}

\maketitle

\section{Introduction}
% In the rapidly evolving landscape of online recruitment, accurate job classification remains a pivotal task for enhancing job recommendation systems, search rankings, and labor market analysis. As job markets expand and diversify, the complexity of job titles and descriptions presents significant challenges in classifying job postings into predefined occupational categories. Traditional text classification techniques often fall short due to the brevity and specificity of job titles, necessitating more sophisticated models that can leverage the intricate relationships inherent in job data.

For online recruitment industry, accurate job classification is crucial, 
as it enhances many services associated with various online job portals---
examples of such services include job recommendation 
system~\cite{cui2021workshop, yang2017combining, zhang2014research}, 
improves search ranking~\cite{li2016get}, and labor market 
analyses~\cite{mortensen1986job}. Accurate job classification also benefits 
job seekers and employers; better matching between job seekers and jobs 
increases both the efficiency and the precision of the recruitment process.

With continuous expansion and diversification of job market, effective 
classification of job postings is becoming increasingly difficult with
many challenges. First, the number of categories into which a job posting
is to be classified is generally large. Besides, the number of jobs
belonging to various categories follow a power-law distribution, making
this a severely imbalanced classification task. Finally, job categories
are arranged into hierarchies with different levels; and job classification
models should be able to solve the classification task with consistent
hierarchical labeling. Traditional hand-picked feature-based 
classification models fail to overcome these challenges, So, it is of 
paramount importance to solve this classification task with novel classification models, which are cognizant of the above challenges.

There exist several published works on job classification 
work~\cite{strah2022job, wang2019deepCarotene, gaspar2020classification}; 
most of these use feature representation of job title, description, or both,
followed by traditional text classification methods. These methods do not
fully leverage the hierarchical nature of job categories, failing to
achieve job embedding vectors retaining hierarchical organization of
job categories. While some efforts~\cite{He2024HierarchicalQC,
wehrmann2018hierarchical} have incorporated hierarchical taxonomies, 
%like the SOC~\footnote{SOC stands for Standard Occupation Classification, developed by US Bureau of Labor Statistics}, 
but challenges remain in effectively modeling the relationships between different levels of job taxonomy nodes. Previous methods~\cite{banerjee-etal-2019-hierarchical, wehrmann2018hierarchical} often utilized linear classifiers or basic hierarchical models, which struggle to capture the complex dependencies between categories and the subtleties inherent in short job titles. These limitations lead to less accurate classifications, particularly when dealing with sparse or noisy training data.

%In recent years, approaches based on representation learning with 
%supervision have gained enormous popularity due to their ability to 
%represent entities in a latent space to support complex classification 
%task. For job classification task, such models can also be used, which
%would leverage semantic features from job titles and descriptions to
%embed jobs in a latent space~\cite{}. Embedding hierarchical job categories into a shared latent space helps tackle imbalanced classification issue, when
%a certain job class does not have enough data instances for robust 
%training. Besides job classification, job embedding also helps in
%various downstream tasks in the only recruitment portals, including
%job recommendation, dynamic matching or ranking of candidates for job
%opportunities as they arise.
%
%By accurately assessing similarity scores between applicants and job categories, this approach facilitates personalized recommendations and ensures that job seekers are presented with the most relevant opportunities. Overall, this integrated method provides a robust framework for improving job classification accuracy and supports more informed decision-making in the recruitment industry.

To address these challenges, we, at X Company\footnote{We are replacing the name of our affiliation with X Company to preserve the anonymity during the review cycle}, introduce a novel 
representation learning and classification model that embeds jobs, and
the nodes of a job taxonomy tree, into a unified lower-dimensional latent
space. Our approach leverages a pre-trained language model, allowing it to
capture the semantic and syntactic details of job data. Meanwhile, joint
embedding of job and job taxonomy nodes makes a taxonomy-aware job embedding,
resulting superior job classification performance. We also utilize X Company's job transition data to overlay a similarity graph over the job taxonomy tree, which is also synthesized for solving this representation
learning task. In summary, our work has the following three key contributions:

    $\bullet$ \textbf{Hierarchy and Similarity-Aware Embeddings}: We develop 
    a hierarchy-aware job embedding model, which embeds both job and job
    category nodes in a unified latent space. Our proposed approach is generic and it can be applied for any data where data instances can be
    classified into nodes of a class taxonomy. This embedding model also
    allows incorporation of node similarity graph for superior embedding
    of entities (jobs) and their class labels (job taxonomy tree nodes).
    
    $\bullet$ \textbf{Innovative Loss Function}: We propose novel loss functions to learn embedding vectors that capture hierarchical job 
    taxonomy and taxonomy node similarity. We validate the effectiveness of these embedding vectors using triplet ranking accuracy (TRA) metric and visualize them in a two-dimensional plot, demonstrating that the taxonomy node embedding vectors successfully preserve both hierarchy and similarity relationships.
    
    $\bullet$ \textbf{Simultaneous Classification}: Our method can simultaneously classify a job into its correct job node class at different levels of taxonomy tree, outperforming existing classification models.

%Hierarchical industry categories, such as those defined by the Standard Occupational Classification (SOC) System published by the United States Department of Labor, provide a structured framework essential for accurate job classification. The SOC system encompasses 23 major groups of job titles and subcategories, forming a comprehensive taxonomy that reflects the intricacies of the job market.
%Our research extends this SOC major system by incorporating an in-house hierarchical taxonomy, Carotene. We propose a model that learns the hierarchical relationships between SOC major groups and Carotene subgroups, enhancing performance at the major level classification while maintaining intra-level relations between the Carotenes. This approach captures both graph and hierarchical relationships, ensuring accurate classification of jobs into their corresponding SOCs and Carotenes.

%In this paper, we detail the development and implementation of our hierarchical job classification model, demonstrating its effectiveness through extensive experimentation on a large-scale dataset of job postings. Our results highlight the model's superior ability to leverage both the hierarchical structure and rich semantic features of job titles and descriptions, significantly improving classification accuracy compared to existing methods.

\section{Related Work} 

Amidst the competitive talent acquisition landscape, a wealth of research has emerged, addressing diverse facets such as the design of job recommendation systems~\cite{cui2021workshop, yang2017combining, zhang2014research}, job-candidate matching~\cite{zhao2021embedding, guo2016resumatcher, ramanath2018towards}, and the creation of shared latent spaces for learning job and skill representation and recommendation~\cite{dave2018combined, al2024interactive, gaspar2020automated}. Further efforts have been devoted to job classification~\cite{strah2022job, wang2019deepCarotene, gaspar2020classification} and query sense disambiguation aimed at enhancing job search~\cite{korayem2015query}.

Hierarchical Text Classification (HTC) can generally be divided into local and global approaches, depending on how the label hierarchy is treated~\cite{zhou-etal-2020-hierarchy}. Local approaches build classifiers for each node or level within the hierarchy, while global approaches build a single classifier for the entire graph. For example, Banerjee et al.\cite{banerjee-etal-2019-hierarchical} developed one classifier per label and transferred parameters from parent models to child models, while Wehrmann et al.\cite{wehrmann2018hierarchical} proposed a hybrid model that combines local and global optimizations. Shimura et al.~\cite{shimura-etal-2018-hft} applied CNNs to leverage data from upper levels to aid categorization in lower levels.

Hierarchical job classification involves assigning job postings to a predefined taxonomy to understand job intent and facilitate downstream recommendation tasks. Traditional approaches for hierarchical job classification, which can be extended to job postings, typically use either a single flattened multi-class classifier or multiple binary classifiers. Feature extraction methods in these approaches can be broadly divided into two categories: N-gram-based~\cite{n-gram-2001} and embedding-based ~\cite{Devlin2019BERTPO}. N-gram-based features rely on the frequency of keywords indicative of the category, while embedding-based features leverage advancements in deep learning and natural language processing to represent queries through word embeddings, like Glove~\cite{pennington-etal-2014-glove} and contextualized embeddings, such as BERT~\cite{Devlin2019BERTPO}, Sentence Transformer~\cite{reimers-2019-sentence-bert}, and Fasttext representation~\cite{bojanowski2017enriching}. For instance, Liu et al.~\cite{Liu2019SystemDO} combined conventional neural networks with Naive Bayes for classification, and Wang et al.~\cite{wang-etal-2022-incorporating} integrated label hierarchy into text encoder through a graph encoder. Additional strategies include using context-aware session information~\cite{context-2019} and searcher engagement data~\cite{HierCat-2023} to enhance classifiers.
Among the other methods, Hi-AGM~\cite{zhou-etal-2020-hierarchy} learns hierarchy-aware label embeddings through the hierarchy encoder and conducts inductive fusion of label-aware text features. HQC ~\cite{He2024HierarchicalQC} designed a modified BCE loss to incorporate the hierarchical relationship among the labels, and HPT~\cite{wang-etal-2022-hpt} injects dependency relations inside the text to classify token labels for hierarchical text classification.

None of the mentioned research works tackle the similarity relation among the taxonomy nodes. To illustrate the similarity relation, we need to consider jobs and taxonomy classes as graph nodes; then the job classification task becomes linking a job node with taxonomy nodes at different levels. Thus the problem transforms into a link prediction task ~\cite{JIN2024128250, kumar2020link, lichtenwalter2010new, zhang2018link}. To tackle the hierarchical and similarity relation, there are multiple papers which incorporate similar kind of relation. For instance, there is a research work which can deal with job and skill relationship and learn the representation of job and skill in the latent space~\cite{job-skill-1, job-skill-2}. We draw upon the link prediction concepts from these studies to design an innovative loss function, enabling a better understanding of taxonomy nodes and enhancing the accuracy of job classification.

\begin{figure}[t]
    \centering
    \includegraphics[scale=0.4]{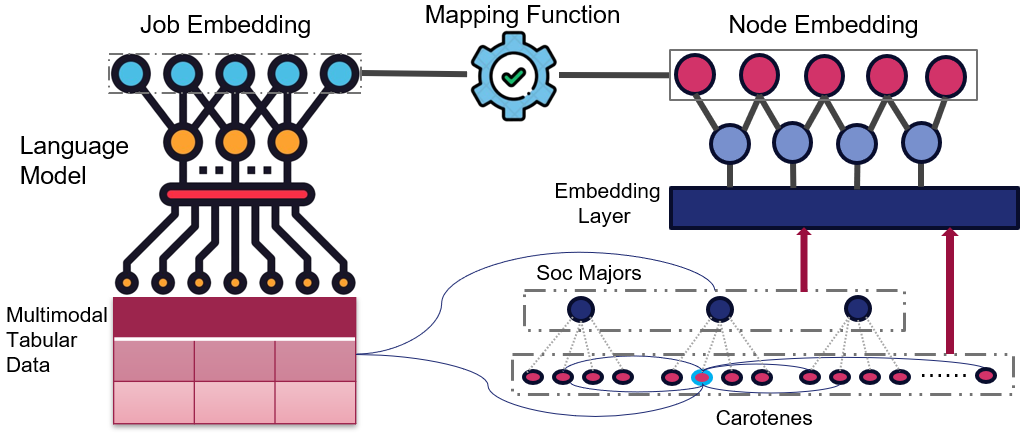}
    \caption{Framework of the Hierarchical Job Classification}
    \label{fig:framework}
\end{figure}

\section{Problem Formulation}
In this section, we provide the mathematical definition of the hierarchical job classification problem. In our dataset the job taxonomy has only two levels, so our discussion will be confined to a two-level taxonomy tree. Nevertheless, the proposed method is extensible to accommodate a multi-level category hierarchy, and graph node classification problem.

\subsection{Notations} 
We first introduce the notations that we use in this paper. Throughout the paper, scalars, different text features, numerical features etc are denoted by lowercase alphabets (e.g., $n$).
Vectors are represented by boldface lowercase letters (e.g., $\mathbf{x}$). The
transpose of the vector $\mathbf{x}$ is denoted by $\mathbf{x}^T$. Bold uppercase letter
(e.g., $\mathbf{X}$) denotes a matrix, and the $i$'th column of a matrix $\mathbf{X}$ is denoted as $\mathbf{x_i}$
. The dot product of two vectors is denoted by $\mathbf{<a,b>}$. The $l_2$ norm of a vector is specified as $|\bf{a}|$ Additionally, we use the uppercase letters for vertex-set, edge-set, and the notation of graph. For the cardinality of a graph we use the absolute symbol. For instance $|E|$ denotes the number of edges in a graph.
\subsection{Objectives}
Say, we are given a set of job postings, $\mathcal{J} = \{j_1, j_2, \ldots, j_N\}$, where each $j_i$ represents the $i^{th}$ job posting. Each job posting is characterized by its title, job description, location, salary, and other relevant features. These features can be collectively represented as $j_i = (t_i, \text{desc}_i, \text{loc}_i, \text{sal}_i, \ldots)$, where $t_i$ denotes the job title, $\text{desc}_i$ represents the job description, $\text{loc}_i$ indicates the job location, $\text{sal}_i$ signifies the salary, and the ellipsis $(\ldots)$ signifies other potential features such as required qualifications, company name, and job type. Each job is assigned a category (SOC) from the set \( S = \{ s_1, s_2, s_3, \ldots ,s_m\} \) and a sub-category (Carotene) from the set \( C = \{ c_1, c_2, c_3, \ldots ,c_n\} \). For instance, \( j_i \) has its SOC label \( s_j \) from \( S \) and the same job has its Carotene \( c_k \) from \( C \). Additionally, each job Carotene \( c_k \) belongs to a SOC \( s_j \). SOC stands for 
Standard Occupation Classification, developed by US Bureau of Labor Statistics. For each SOC node, there is a collection of Carotene
nodes, which are finer classification of each SOC node developed in-house by X Company. Say, a SOC node \( s_j \)  is associated with a few Carotenes \( C_s = \{ c_m, c_n, c_o, \ldots \} \) where \( C_s \subset C \). Additionally, there is a directed similarity graph \( G = (V, E) \) among the Carotenes where \( |V| = |C| \), and any two Carotenes \( c_p \) and \( c_q \) can have an edge \( \delta \) from \( c_p \) to \( c_q \) such that \( \delta \in E \).

The goal is to leverage the information to learn a function $\mathcal{F}$ that maps a job posting \( j_i \) to its corresponding SOC label \( s_j \) and Carotene label \( c_k \), such that $\mathcal{F}(j_i) \rightarrow (s_j, c_k)$, where $j_i \in s_j$ and $j_i \in c_k$. Furthermore, we want to represent the job, SOC, and Carotene embeddings in such a way that both the hierarchy and graph relationships among the nodes are captured. Specifically, we aim to learn embeddings \( \mathbf{E_j} \), \( \mathbf{E_s} \), and \( \mathbf{E_c} \) for jobs, SOCs, and Carotenes, respectively, such that: the embeddings reflect the hierarchical relationship: 
$d(\mathbf{e_{j_i}, e_{s_j}}) < d(\mathbf{e_{j_i}, e_{s_k})}$ for $s_j = \text{SOC}(j_i)$, $s_k \neq s_j$, 
    $d(\mathbf{e_{j_i}, e_{c_k}}) < d(\mathbf{e_{j_i}, e_{c_m}})$ for $c_k = \text{Carotene}(j_i)$, $c_m \neq c_k$ 
    and $d(\mathbf{e_{s_j}, e_{c_k}}) < d(\mathbf{e_{s_j}, e_{c_t}})$ for $c_k  \in \text{children}(s_j)$, $c_t \notin \text{children}(s_j)$
    where \( d(\cdot, \cdot) \) denotes a distance metric.
    Additionally, another goal is-- the embeddings capture the graph structure among Carotenes:
$d(\mathbf{e_{c_p}, e_{c_q}}) < d(\mathbf{e_{c_p}, e_{c_r}})$ for $(c_p, c_q) \in E$, $(c_p, c_r) \notin E.$

Figure~\ref{fig:framework} illustrates the overall problem and framework of the hierarchical classification challenge. As shown in the figure, the job data consists of multimodal tabular data, which is first processed by a language model to understand its content. The SOCs and Carotenes maintain a hierarchical relationship, as depicted. Additionally, the Carotenes are represented in a similarity graph where any two Carotenes can be similar, regardless of whether they share a common parent. The mapping function $\mathcal{F}$ is responsible for assigning each job to its corresponding SOC and Carotene. Furthermore, we need to learn the embeddings of the job, SOC, and Carotene in such a way that both the hierarchical and graph relationships among the nodes are captured by the node representation vectors.

To summarize, the learned function $\mathcal{F}$ ensures that job postings are correctly classified into their respective SOCs and Carotenes while preserving the hierarchical and similarity relationships inherent in the data.
\section{Methodology}

In this section, we present our novel representation learning and classification model designed to embed jobs, SOCs, and Carotenes into a k-dimensional latent embedding space. This innovative approach captures both graph and hierarchical relationships, ensuring accurate classification of jobs into corresponding SOCs and Carotenes.

\subsection{Job Embedding}

Attention-based transformer models excel at converting job postings into embeddings by effectively capturing long-range dependencies and contextual information. Using self-attention mechanisms, these models weigh the relevance of different words, creating rich, context-aware embeddings that accurately represent the detailed information in job postings. This is essential for handling the diverse content on X Company, where the model must support parallel computation while maintaining deep contextual understanding. To achieve this, we transform each job posting \( j_i \)
into an embedding using a state-of-the-art pretrained language model, such as BERT, or Sentence Transformers. 
First, all the features of $j_i$ are concatenated as the following:

\[\hat{j_i} = t_i + \text{desc}_i + \text{loc}_i + \textFunction{\text{sal}_i} + , \ldots \]
where $\textFunction{c}$, and \textbf{+} denote casting a value into text and string-level concatenation operation respectively. A tokenizer is then applied to tokenize $\hat{j_i}$ as the following:

\[
W = [w_1, w_2, \ldots, w_n]
\]
where \( w_k \) represents the \( k \)-th token in $\hat{j_i}$. This sequence of tokens is then passed through the transformer model, which applies multiple layers of self-attention and feed-forward neural networks. The self-attention mechanism calculates the attention scores for each token pair.
First the query (\(\mathbf{Q}\)), key (\(\mathbf{K}\)), and value (\(\mathbf{V}\)) matrices are calculated:

\[
\mathbf{Q} = \mathbf{W_Q} \mathbf{X}, \quad \mathbf{K} = \mathbf{W_K} \mathbf{X}, \quad \mathbf{V} = \mathbf{W_V} \mathbf{X}
\]
where \(\mathbf{X}\) is the input token embeddings matrix, and \(\mathbf{W_Q}\), \(\mathbf{W_K}\), \(\mathbf{W_V}\) are learned weight matrices.
Then the attention scores are calculated using the scaled dot-product attention:
\[
\text{Attention}(\mathbf{Q}, \mathbf{K}, \mathbf{V}) = \text{softmax}\left(\frac{\mathbf{Q} \mathbf{K}^T}{\sqrt{d_k}}\right) \mathbf{V}
\]
where \( d_k \) is the dimensionality of the key vectors.
The weighted sum of values are aggregated to obtain the contextualized token representations:
\[
\mathbf{Z} = \text{Attention}(\mathbf{Q}, \mathbf{K}, \mathbf{V})
\]
These contextualized representations are then passed through several transformer layers, each consisting of self-attention and feed-forward sub-layers, to capture deeper semantic relationships.

The final layer's output is a sequence of embeddings for the tokens, which is typically pooled to obtain a single embedding vector representing the entire job posting \( j_i \):

\[
\mathbf{e_{j_i}} = \text{Pooling}(\mathbf{Z})
\]
where the pooling operation can be the [CLS] token output, aggregation over the token embeddings. The resulting embedding vector \( \mathbf{e_{j_i}} \) resides within a $q$ dimensional latent space,  effectively capturing the semantic nuances and contextual information of the job description.

\subsection{SOC and Carotene Embedding}
Each SOC and Carotene is linked to specific signature texts, which act as target labels for the hierarchical classification task. To effectively capture the hierarchical structure and similarity relationships, it is essential that the representations of SOCs and Carotenes be trainable—this is highlighted as the first major contribution of our research. While a transformer-based approach for text representation could provide a more sophisticated model, it would also introduce unnecessary complexity and increase processing time. Therefore, we opted for a simpler, trainable single-layer representation for SOCs and Carotenes.

We have two different trainable matrices $\mathbf{E_s}$, and $\mathbf{E_c}$ where $\mathbf{E_s} = \{\mathbf{e_{s_1}^T}, \mathbf{e_{s_2}^T}, ... , \mathbf{e_{s_m}^T}\} \in \mathbf{\mathbb{R}}^{m \times q}$ , and $\mathbf{E_c} = \{\mathbf{e_{c_1}^T}, \mathbf{e_{c_2}^T}, ... , \mathbf{e_{c_n}^T}\} \in \mathbf{\mathbb{R}}^{n \times q}$. $\mathbf{e_{s_j}^T}$ and $\mathbf{e_{c_k}^T}$ represents the $j$'th and $k$'th columns of matrices $\mathbf{E_s}$ and $\mathbf{E_c}$ respectively. The embedding of SOC $s_j$, and Carotene $c_k$ are respectively $\mathbf{e_{s_j}^T}$ and $\mathbf{e_{c_k}^T}$. Both the matrices $\mathbf{E_s}$, and $\mathbf{E_c}$ should preserve the connectivity relations among job-soc, soc-carotene, and carotene-carotene. The dimensionality of the SOC and Carotene embeddings is set to $q$, matching the dimensionality of the word embeddings from the language model. As we have text embedding coming from a language model, $\mathbf{E_s}$, and $\mathbf{E_c}$ matrices should learn from fusion of knowledge from text data, hierarchical, and graph relations.

\subsection{Loss Components}

Since hierarchical text classification is a multi-label classification problem, previous methods \cite{zhou-etal-2020-hierarchy, wang-etal-2022-hpt} primarily treat HTC as a series of multiple binary classification tasks. Consequently, they employ binary cross-entropy (BCE) as their loss function, which is defined as follows:

\[
\mathcal{L}_{\text{BCE}} = - \sum_{i \in C} \left[ y_i \log(s_{y_i}) + (1 - y_i) \log(1 - s_{y_i}) \right]
\]
where \( C \) represents the set of all classes, \( y_i \) is the ground truth label for class \( i \), and \( s_{y_i} \) is the predicted probability for class \( i \). This loss can classify all multi-label hierarchical labels. 

However, BCE loss and its variants are not well-suited for working with the Carotene similarity graph. This led us to develop a simpler multi-level loss function. We created a categorical cross-entropy loss function for both the SOC and Carotene levels. Since we aim to classify a job into its correct SOC and Carotene simultaneously (as illustrated as our third major contribution), we designed two multi-class classification loss functions based on the same job embedding.

The representation of the job $j_i$, $\mathbf{ e_{j_i} }$ is passed through a classifier layer for SOC classification, \( s_j \) using a categorical cross-entropy loss:

\[
\mathcal{L}_{\text{soc}} = - \sum_{i=1}^{N} \sum_{j=1}^{m} y_{ij} \log ( \sigma (\mathbf{e_{j_i}} \cdot \mathbf{W}_{soc}))
\]

where \( y_{ij} \) is the true label indicating whether job \( j_i \) belongs to SOC \( s_j \), \( \sigma \) is the softmax function, and $\mathbf{W}_{soc}$ is the trainable weight matrix for SOC classification.

Similarly, the job embedding $\mathbf{ e_{j_i} }$  is classified into the corresponding Carotene \( c_k \) using another categorical cross-entropy loss:

\[
\mathcal{L}_{\text{carotene}} = - \sum_{i=1}^{N} \sum_{k=1}^{n} y_{ik} \log \left( \sigma (\mathbf{e_{j_i}} \cdot \mathbf{W}_{carotene}) \right)
\]
where \( y_{ik} \) is the true label indicating whether job \( j_i \) belongs to Carotene \( c_k \).

To capture the comprehensive information in the context of hierarchical relationship, in the literature there are multiple contrastive learning methods. Let assume the triplet \( (s_j, c_k, c_t) \) where \( c_k \) directly linked to \( s_j \) and \( c_t \) not linked to \( s_j \). There are methods~\cite{He2024HierarchicalQC} which use contrastive loss function --

\[
\mathcal{L}_{\text{contrastive}} = -log \frac{exp(<\mathbf{e}_{s_j}, \mathbf{e}_{c_t}>)}{\sum(exp(<\mathbf{e}_{s_j},\mathbf{e}_{c_k}>)} 
\] without margin. As it is easier to separate the  
positive node than the negative one using the contrastive loss above, and the margin determines the level of separability we use a margin loss instead 
such that the embedding of a Carotene directly linked to a SOC is more similar to the SOC embedding than other Carotenes:

\[
\mathcal{L}_{\text{soc-car}} = \sum_{(s_j, c_k, c_t)} \left[ \alpha + \frac{<\mathbf{e}_{s_j}, \mathbf{e}_{c_k}>}{|\mathbf{e}_{s_j} ||\mathbf{e}_{c_k}|} -   \frac{<\mathbf{e}_{s_j}, \mathbf{e}_{c_t}>}{|\mathbf{e}_{s_j} ||\mathbf{e}_{c_t}|}
 \right]_+
\]
where \( (s_j, c_k, c_t) \) are triplets with \( c_k \) directly linked to \( s_j \) and \( c_t \) not linked to \( s_j \), \( \alpha \) is a margin, and \([ \cdot ]_+\) denotes the hinge loss. The designed loss function aims to capture the hierarchy relation between SOCs and Carotenes.

To maintain the similarity graph structure among Carotenes, we define another margin loss such that Carotenes with a direct edge have higher similarity:

\[
\mathcal{L}_{\text{car-car}} = \sum_{(c_p, c_q, c_r)} \left[ \alpha + \frac{<\mathbf{e}_{c_p}, \mathbf{e}_{c_q}>}{|\mathbf{e}_{c_p} ||\mathbf{e}_{c_q}|} -   \frac{<\mathbf{e}_{c_p}, \mathbf{e}_{c_r}>}{|\mathbf{e}_{c_p} ||\mathbf{e}_{c_r}|}
 \right]_+
\]
where \( (c_p, c_q, c_r) \) are triplets with \( (c_p, c_q) \in E \) and \( (c_p, c_r) \notin E \).

To integrate the hierarchical and similarity relationships with the contextual information from the job embedding, we define two additional loss components: $\mathcal{L}_{\text{job-soc}}$ and $\mathcal{L}_{\text{job-carotene}}$ for this purpose.

\[
\mathcal{L}_{\text{job-soc}} = \sum_{(j_i, s_j, s_k)} \left[ \alpha + \frac{<\mathbf{e}_{j_i}, \mathbf{e}_{s_j}>}{|\mathbf{e}_{j_i} ||\mathbf{e}_{s_j}|} -   \frac{<\mathbf{e}_{j_i}, \mathbf{e}_{s_k}>}{|\mathbf{e}_{j_i} ||\mathbf{e}_{s_k}|}
 \right]_+
\] 
where \( (j_i, s_j, s_k) \) are triplets with \( (j_i, s_j) \in E \) and \( (j_i, s_k) \notin E \) and similarly 
$\mathcal{L}_{\text{job-car}}$ is designed as the following where \( (j_i, c_j, c_t) \) are triplets with \( (j_i, c_j) \in E \) and \( (j_i, c_t) \notin E \)-- 

\[
\mathcal{L}_{\text{job-car}} = \sum_{(j_i, c_j, c_t)} \left[ \alpha + \frac{<\mathbf{e}_{j_i}, \mathbf{e}_{c_j}>}{|\mathbf{e}_{j_i} ||\mathbf{e}_{c_j}|} -   \frac{<\mathbf{e}_{j_i}, \mathbf{e}_{c_t}>}{|\mathbf{e}_{j_i} ||\mathbf{e}_{c_t}|}
 \right]_+
\] 

The final loss function is a weighted sum of all the loss components:

\[
\begin{aligned}
\mathcal{L}_{\text{total}} = &\ \lambda_{\text{1}} \mathcal{L}_{\text{soc}} + \lambda_{\text{2}} \mathcal{L}_{\text{carotene}} + \lambda_{\text{3}} \mathcal{L}_{\text{soc-car}} \\
&+ \lambda_{\text{4}} \mathcal{L}_{\text{car-car}} +  \lambda_{\text{5}} \mathcal{L}_{\text{job-soc}} +  \lambda_{\text{6}} \mathcal{L}_{\text{job-car}}
\end{aligned}
\]

where \( \lambda_{\text{1}}, \lambda_{\text{2}}, \lambda_{\text{3}}, \lambda_{\text{4}}, \lambda_{\text{5}}, \),  and \( \lambda_{\text{6}} \) are hyperparameters that control the contribution of each loss component to the total loss, and all of these are in the range (0,1).
The goal of the framework is to ensure that job postings are correctly classified into their respective SOCs and Carotenes while preserving the hierarchical and similarity relationships inherent in the data.

%%%%%  TRA = Triplet Rangking Accuracy
%%% HQC = Hierarchical Query Classification in E-commerce Search
%%%% HITIN =  Hierarchy-aware global model for hierarchical text classification (2020)
%%% HPT = Hierarchy-aware prompt tuning for hierarchical text classification (2022)

\section{Experiments and Results}

\begin{table*}[h]
    \caption{Performance of all experimental methods; hyphen indicates no experiment result.}  
    \setlength{\tabcolsep}{1.5pt} % Reduce column separation
    \centering
    \renewcommand{\arraystretch}{1.1} % Slightly reduce row height
    \scalebox{0.9}{ % Scale down the entire table
        \begin{tabular}{l | c | c| c| c | c | c}
            \hline
            \bf Method & \bf Acc (SOC) & \bf Acc (Car) & \bf Prec (Car)  &  \bf Recall (Car) & \bf TRA (soc-car) & \bf TRA (car-car) \\
            \hline
            BERT (v1) & 0.867  & 0.855 & 0.842 & 0.840 & - & - \\
            BERT (v2) & 0.883  & 0.852 & 0.845 & 0.842 & - & - \\
            Sentence-Transformer (v1) & 0.869 & 0.852 & 0.841 & 0.839 & - & - \\
            Sentence-Transformer (v2) & 0.886 & 0.853 & 0.846 & 0.843 & - & - \\
            \hline
            HQC & 0.887  & 0.863 & 0.849 & 0.847 & 0.896 & - \\
            HiAGM & 0.403 & 0.332 & 0.320 & 0.325 & 0.730 & - \\
            HPT & 0.882 & 0.853 & 0.847 & 0.842 & 0.901 & - \\
            \hline
            Hierarchical-Classification (Soft) &  0.932 &  0.877 & 0.851 & 0.849 &  \bf 0.999 & - \\
            Hierarchical-Classification (Hard) &  0.942 &  0.879 & 0.852 & 0.852 &  \bf 0.999 & - \\
            Hierarchical-Sim-Classification (Soft) &  0.937 & 0.877 & 0.849 & 0.847 & \bf 0.999 & \bf 0.999 \\
            \bf Hierarchical-Sim-Classification (Hard) &  \bf 0.948 & \bf 0.893 & \bf 0.856 & \bf 0.855 & 0.998 &  0.998 \\
            Hierarchical-Sim-Classification (Text) &  0.939 & 0.877 & 0.852 & 0.849 &  0.995 &  0.998 \\
            \hline
        \end{tabular}
    }
    \label{table:experiments}
\end{table*}

\subsection{Triplet Generation Technique}
In the previous section, we mentioned four loss components $\mathcal{L}_{\text{soc-car}}$, $\mathcal{L}_{\text{job-soc}}$, $\mathcal{L}_{\text{job-car}}$, and $\mathcal{L}_{\text{car-car}}$. Among these, $\mathcal{L}_{\text{soc-car}}$ works on the triplets for hierarchical relation. For instance, let \( (s_j, c_k, c_t) \) be a triplet where \( c_k \) is directly linked to \( s_j \) and \( c_t \) is not linked to \( s_j \). For all the variants of \name, whenever there is an edge between a SOC and a Carotene, we find the $n\_neg$ number of Carotenes which are not children of the particular SOC. All these triplets are then saved to a file, and we have another hyperparameter $n\_sample$ to randomly sample from all the triplets in a batch. Similarly, for the Carotene-to-Carotene similarity graph, we generate $n\_neg$ number of triplets for every edge and save them in a file. The exact same $n\_sample$  hyperparameter is used to sample from the list of triplet files. Among the variants of \name, Hierarchical Classification (Soft) works with only soc-carotene triplets, and Hierarchical Sim Classification (Soft) works with both soc-carotene and carotene-carotene triplets. For the Hierarchical-Classification (Hard) method, we choose which Carotenes need to be selected as negative triplets. For that, we first select the Carotenes which are not children of the SOC, and then rank the top $n\_neg$ Carotenes which have the best cosine similarity with the SOC signature text embedded by the sentence-transformer. For the Hierarchical Sim Classification (Hard) method, we find the hard triplets for both soc-carotene and carotene-carotene in a similar fashion. For the $\mathcal{L}{\text{job-soc}}$ and $\mathcal{L}{\text{job-car}}$ loss components, job-soc and job-carotene triplets are generated and sampled in a manner consistent with the approach used for other edges.

\subsection{Dataset Description} Our dataset is derived from our internal X Company database, containing millions of job listings. We filtered the dataset to include only active jobs. The features available for use include location, job type, salary, job description, and job title, among others. For selecting Carotenes, we considered only the job title and job description as these are the most relevant features. The master dataset comprises 456K instances, which are split into training, testing, and validation sets maintaining 6:2:2 ratio randomly. The number of instances in train, test, and validation splits are 273K,91K, and 91K respectively. Each job entry is labeled with a SOC and a Carotene. Additionally, a hierarchical tree structure illustrates the relationships between SOCs and their child Carotenes. Our dataset includes 23 SOCs and 4,778 Carotenes in total, with each SOC connected to an average of 210 Carotene nodes. The SOCs have a minimum degree of 67 and a maximum degree of 495.

Moreover, we have a Carotene similarity graph indicating which Carotenes are similar to each other. This is an directed graph based on similarity scores between Carotenes. To manage complexity, we enforce an internal threshold and ranking criteria to ensure that each Carotene is connected to a maximum of five other Carotenes, resulting in a maximum degree of five in the carotene-carotene similarity graph. Finally, for each SOC and Carotene, we have signature texts that describe their functions.

\begin{figure}[t]
    \centering
    \includegraphics[scale=0.28]{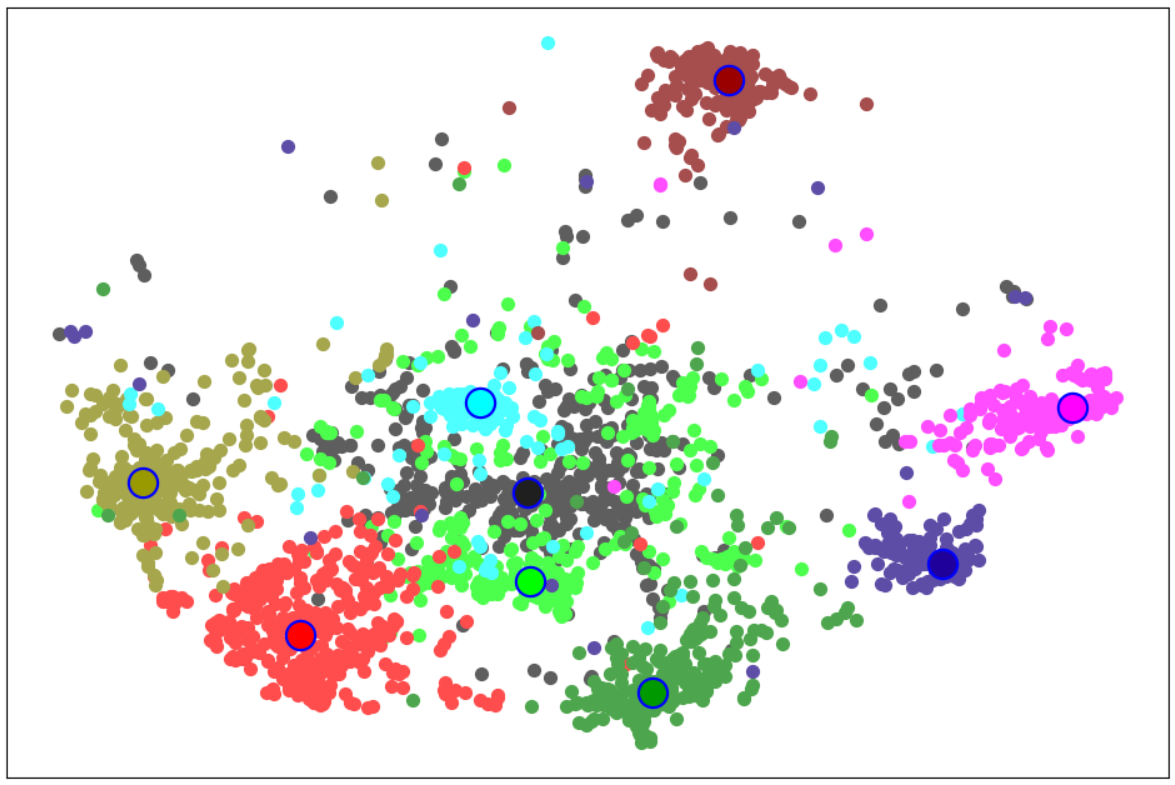}
    \caption{t-SNE Plot of SOCs and Their Corresponding Carotene Embeddings from Hierarchical-Sim-Classification (Hard)}
    \label{fig:embedding}
\end{figure}

\subsection{Description of the Competing Methods} 
We have implemented two classification models using \\\ BERT~\cite{Devlin2019BERTPO} and Sentence Transformers~\cite{reimers-2019-sentence-bert}, each with two versions. Version 1 focuses primarily on Carotene classification, without incorporating SOC classification into the loss function. Version 2 averages the SOC cross-entropy and Carotene cross-entropy losses, with the weights being tunable hyper-parameters.

We evaluated three models for hierarchical text classification: HQC~\cite{He2024HierarchicalQC}, HiAGM~\cite{zhou-etal-2020-hierarchy}, and HPT~\cite{wang-etal-2022-hpt}. "Hierarchical Query Classification in E-commerce Search" (HQC) designed a modified BCE loss to incorporate the hierarchical relationship among the labels. The best version of the "Hierarchy-aware Global Model for Hierarchical Text Classification" (HiAGM) captures label information from GCN and combines it with text embeddings from an attention-based approach, using BCE loss for hierarchical classification. "Hierarchy-aware Prompt Tuning for Hierarchical Text Classification" (HPT) injects dependency relations inside the text to classify token labels for hierarchical text classification. However, none of these baselines can adapt to the similarity transition graph.

Among our designed methods, we have multiple versions of the Hierarchical Classification model. Hierarchical Classification (Soft) works with randomly selected soc-carotene triplets, without considering hard negative Carotenes relative to the signature text. In contrast, Hierarchical Classification (Hard) uses intelligently selected hard negatives. Similarly, the Hierarchical Sim Classification (Hard) method handles hard negative triplets for both soc-carotene and carotene-carotene edges. Another variant, Hierarchical Sim Classification (Text), is a different version of Hierarchical-Sim-Classification (Hard), working on hard negative triplets for soc-carotene and carotene-carotene edges. The key difference is the use of two pre-trained SOC embedding and Carotene pre-trained layers, with weights derived from the corresponding SOC and Carotene signature texts.

\begin{table*}
   \caption{Ablation Study with Hierarchical Classification (Hard) Method }  
   \setlength{\tabcolsep}{1.9pt}
        \centering
            \renewcommand{\arraystretch}{1.1}
            \scalebox{1}{
                \begin{tabular}{l | c | c | c | c | c| c | c|c|c}
                    %\multicolumn{4}{c}{\bf Hyponym-Hypernym Classification Comparison} \\
                    \hline
                    \bf $ \boldsymbol{\lambda}_1 $ & \bf $ \boldsymbol{\lambda}_2 $ & \bf $ \boldsymbol{\lambda}_3 $ & \bf $ \boldsymbol{\lambda}_4 $ & \bf $ \boldsymbol{\lambda}_5 $ & \bf $ \boldsymbol{\lambda}_6 $ &\bf Acc (SOC) & \bf Acc (Carotene) & \bf TRA (soc-carotene) & \bf TRA (carotene-carotene) \\\hline
                     0.0 & 0.7 & 0.3 & 0.3 & 0.3 & 0.3 & 0.331 & 0.873 & 0.998 & 0.999 \\
                     0.3 & 0.0 & 0.3 & 0.3 & 0.3 & 0.3 & 0.953  & 0.02 & 0.998 & 0.998 \\
                     0.3 & 0.7 & 0.0 & 0.3 & 0.3 & 0.3 & 0.934 & 0.884 & 0.694 & 0.997 \\
                     0.3 & 0.7 & 0.3 & 0.0 & 0.3 & 0.3 & 0.942 & 0.879 & 0.999 & 0.688 \\
                     0.3 & 0.7 & 0.3 & 0.3 & 0.3 & 0.3 & 0.948 & 0.893 & 0.998 & 0.998 \\
                      \hline
                \end{tabular}
            }
        \label{table:ablation}
\end{table*}

\subsection{Results} We conducted extensive experiments on hierarchical text classification using our X Company dataset.  We evaluated the predictions using accuracy metrics for both SOC and Carotene. Additionally, we introduced a novel evaluation metric, TRA (Triplet Ranking Accuracy), which measures the percentage of times a triplet is correctly ordered based on the derived embeddings and cosine similarity. To compute this metric, we considered all the generated triplets stored in the file. 
Table~\ref{table:experiments} summarizes all experimental results, with hyphens indicating setups where no experiments were conducted.

Among the baseline methods—BERT (v1), BERT (v2), Sentence Transformer (v1), and Sentence-Transformer (v2)—Carotene test accuracy is similar. However, models with a separate categorical cross-entropy for SOC classification (version 2) outperform version 1 models. Among HQC, HiAGM, and HPT, HQC achieves the best classification accuracy at both SOC and Carotene levels. This is attributed to HQC being a transformer-based model with an improved loss function, whereas HiAGM, a non-scalable bi-LSTM based model, struggles with our large-scale dataset of nearly 5,000 labels, an average degree of 210, and 300K job entries. Nonetheless, HPT slightly outperforms HQC in terms of TRA among the baseline methods.

The last category of models in the table represents our custom-designed methods. All of these methods surpass the baseline methods in every evaluation metric. Notably, the Hierarchical Sim Classification (Hard) model outperforms other variants in SOC and Carotene accuracy, achieving a best SOC accuracy of 0.948 and a best Carotene accuracy of 0.893. In terms of TRA, the Hierarchical Sim Classification (Soft) model has slightly better accuracy than the Hierarchical Sim Classification (Hard) model, primarily due to a lower margin and the introduction of more triplets in the loss function.

We evaluated the embeddings from Hierarchical Sim Classification (Hard) using t-SNE with a margin of 0.4. The SOC and Carotene embeddings were first reduced to 2 dimensions to fit into the 2D space. We randomly selected 10 SOCs and their corresponding Carotene embeddings. Each SOC is represented by a large circle of a different color, distinguished by a blue ring. The corresponding Carotene embeddings cluster around their respective SOC embeddings. For example, the red SOC embedding is surrounded by its child Carotene embeddings, most of which are also red. The visualization clearly shows that the learned embeddings tend to cluster around the correct SOC embeddings. Additionally, some Carotenes overlap with others, indicating that the task involves both hierarchical classification and the similarity relationships among Carotenes. The overlapping Carotenes, the clustering tendency, and the same-colored Carotenes surrounding the SOC embeddings demonstrate that the designed loss function effectively captures both hierarchical and similarity relationships among the SOC and Carotene nodes. This comprehensive understanding of node relations allows the Hierarchical Sim Classification (Hard) model to achieve the best accuracy among the baseline methods.

\begin{table*}[h]
    \caption{Predictions of the Jobs by Hierarchical Sim classification (Hard)}  
    \setlength{\tabcolsep}{1.5pt} % Reduce column separation
    \centering
    \renewcommand{\arraystretch}{1.1} % Slightly reduce row height
    \scalebox{0.9}{ % Scale down the entire table
        \begin{tabular}{l | c | c | c | c}
            \hline
            \bf Job Title & \bf Actual SOC & \bf Predicted SOC & \bf Actual Carotene & \bf Predicted Carotene \\\hline
            Port Service Technician - Crane Technician
            & 49 & 49 & Crane Technician & Crane Technician \\
            In House Tradesman
            & 11 & 51 & Tradesman & Craftsman \\
            Specialized Clinical Supervisor
            & 21 & 21 & Clinical Supervisor & Clinical Supervisor \\
            Donor Records Coordinator
            & 27 & 27 & Fundraising Coordinator & Fundraising Coordinator \\
            SBA Credit Associate
            & 13 & 43 & Credit Administrator & Administrative Associate \\
            Travel Dialysis RN
            & 29 & 29 & Dialysis Nurse & Licensed Practical Nurse \\
            \hline
        \end{tabular}
    }
    \label{table:predictions}
\end{table*}

\section{Hyper-Parameter Tuning} Various hyperparameters are associated with all variants of Hierarchical Classification. Specifically, $\lambda_1$, $\lambda_2$, $\lambda_3$, $\lambda_4$, $\lambda_5$ and $\lambda_6$ are included in the loss function outlined in the methodology section. These hyperparameters assign different weights to the various components of the loss function. The $\lambda$ values are tuned within the range of $[0.1, 0.9]$ with an interval of 0.1. We observed the highest SOC and Carotene accuracy when $(\lambda_1, \lambda_2, \lambda_3, \lambda_4, \lambda_5, \lambda_6) = (0.3, 0.5, 0.3, 0.3, 0.3, 0.3)$ for the Hierarchical Sim Classification (Hard) method. These $\lambda$ values are also applicable to other variants, and the same hyperparameter $\lambda$s are used across different models.

The batch size of the model is fixed at 4 for all variants and is not tuned. The number of epochs is set to 100 for all variants, with early stopping enabled and patience set to 3. This means that if the validation accuracy does not improve for three consecutive epochs, the model training stops early. The learning rate is also fixed at $2 \times 10^{-5}$ using the Adam optimizer for all variants.

When creating job-soc triplets, the hyperparameter $n\_neg$ is tuned within the range of $[5, 20]$ with a step of 5, finding that $n\_neg = 15$ yields the best results for Hierarchical Sim Classification (Hard), Hierarchical Sim Classification (Text), and Hierarchical Classification (Hard). For Hierarchical Classification (Soft) and Hierarchical Sim Classification (Soft), $n\_neg = 20$ provides the best results. For job-carotene, soc-carotene, and carotene-carotene triplets, the hyperparameter $n\_neg$ is tuned within the set \{10, 50, 100, 150, 200, 250, 300, 350, 400, 450, 500\}, with $n\_neg = 200$ achieving the best performance across all variants.

For the sampling hyperparameters to sample from the triplets, $n_sample$ is fixed for all variants. Finally, the margin hyperparameter is tuned within the range $[0.1, 0.6]$ with an interval of 0.1. A lower margin increases TRA (soc-carotene) and TRA (carotene-carotene), but does not optimize SOC and Carotene accuracy. Additionally, t-SNE plots fail to produce distinct SOC and Carotene embeddings when the margin is low, resulting in scattered embeddings across the 2D space. We found the best accuracy with a margin of 0.4, as higher margins do not guarantee the best accuracy achieved by the model.

\section{Ablation Study} 

The ablation study, presented in Table \ref{table:ablation}, assesses the impact of each loss component on the performance of our Hierarchical Classification (Hard) method. For all the experiments, we fixed the number of epochs to 100 with early stopping tolerance set to 3, batch size to 4, and margin to 0.4. 
For all the experiments we fix the values of \(\lambda_5\), \(\lambda_6\) to the most optimum values 0.3 to fuse the hierarchy and similarity relation with contextual job information.
When the SOC classification loss component is removed (\(\lambda_1 = 0\)), SOC accuracy significantly drops to 0.331, underscoring its critical role in accurate SOC classification. Conversely, removing the Carotene classification loss (\(\lambda_2 = 0\)) results in a dramatic decrease in Carotene accuracy to 0.02, highlighting its importance for Carotene classification. The soc-carotene contrastive loss (\(\lambda_3\)) is essential for maintaining the hierarchical relationship between SOCs and Carotenes, as indicated by a significant drop in TRA (soc-carotene) to 0.694 when \(\lambda_3\) is set to 0.0. Similarly, the carotene-carotene contrastive loss (\(\lambda_4\)) is crucial for capturing similarity relationships among Carotenes, with TRA (carotene-carotene) dropping to 0.688 without it. The best performance across all metrics, with a SOC accuracy of 0.948 and Carotene accuracy of 0.893, is achieved when all loss components are included (\(\lambda_1=0.3\), \(\lambda_2=0.7\), \(\lambda_3=0.3\), \(\lambda_4=0.3\)), demonstrating a balanced contribution from each component in capturing both hierarchical and similarity relationships within the data.

\section{Case Study}

Table~\ref{table:predictions} highlights selected job titles from the X Company dataset, comparing ground truth and predicted SOC and Carotene labels from the Hierarchical Text Classification (Hard) model, which achieved 95% SOC classification accuracy.

Most predictions align with the ground truth, such as for \texttt{Port Service Technician - Crane Technician''}, \texttt{Specialized Clinical Supervisor''}, and \texttt{Donor Records Coordinator''}. However, the job \texttt{In House Tradesman''} was completely misclassified, with the SOC incorrectly predicted as 51 instead of 11, and Carotene as \texttt{Craftsman''} instead of \texttt{Tradesman''}, despite there being no similarity edge between these categories. This is a minor exception.

Another notable case is \texttt{SBA Credit Associate''}, where the SOC was incorrectly predicted as 43 instead of 13. However, the predicted Carotene, \texttt{Administrative Associate''}, is related to the actual Carotene, \texttt{Credit Administrator''}, by a similarity edge. For \texttt{Travel Dialysis RN''}, the SOC was correctly predicted, but the Carotene was incorrectly identified as \texttt{Licensed Practical Nurse''} instead of \texttt{Dialysis Nurse''}, though these categories are closely related.

\section{Conclusion}
This research introduced a hierarchical job classification model that effectively embeds jobs and industry categories like SOC and Carotene into a shared latent space, capturing both graph and hierarchical relationships. The model surpasses traditional methods in handling complex job market data, particularly improving classification accuracy and mitigating cold start issues. The successful integration of SOC with the Carotene taxonomy underscores the model's robustness, offering a powerful tool for informed decision-making in recruitment. This work advances job classification methodologies and opens avenues for applying similar models to other hierarchical and graph-based classification challenges.

\bibliographystyle{ACM-Reference-Format}
\balance
\bibliography{sample-base}
\end{document}